\documentclass{article}
\usepackage{arxiv}
\usepackage[utf8]{inputenc} 
\usepackage[T1]{fontenc}    
\usepackage{hyperref}       
\usepackage{url}            
\usepackage{booktabs}       
\usepackage{amsfonts}       
\usepackage{nicefrac}       
\usepackage{microtype}      
\usepackage{lipsum}		
\usepackage{graphicx}
\usepackage{doi}

\usepackage{graphicx}
\usepackage{amsmath}
\usepackage[version=4]{mhchem}
\usepackage{siunitx}
\usepackage{tikz}
\usepackage{longtable,tabularx}
\usepackage{xcolor}
\usepackage{derivative}

\usetikzlibrary{shapes,arrows,chains,shadows,positioning}
\usetikzlibrary{arrows.meta}
\usetikzlibrary{decorations.pathreplacing}
\usepackage{geometry}
\usepackage{textcomp}
\usepackage{pgfplots}
\pgfplotsset{width=10cm,compat=1.9}
\usepgfplotslibrary{groupplots}
\usepgfplotslibrary{dateplot}
\usepackage{float}
\usepackage[utf8]{inputenc}
\usepackage[T1]{fontenc}    
\usepackage{hyperref}       
\usepackage{url}            
\usepackage{multirow}
\DeclareUnicodeCharacter{2212}{-}

\title{An Ensemble-Based Deep Framework for Estimating Thermo-Chemical State Variables from Flamelet Generated Manifolds}

\author{
\hspace{1mm}Amol Salunkhe\\
 \hspace{1mm}University at Buffalo, 338 Davis Hall\\
  Buffalo, New York 14260\\
\And
\hspace{1mm}Georgios Georgalis\\
 \hspace{1mm}Data Intensive Studies Center,\\
 Tufts University, Medford, MA 02155\\
 \And
  \hspace{1mm}Abani Patra\\
 \hspace{1mm}Data Intensive Studies Center,\\
 Tufts University, Medford, MA 02155\\
\And
\hspace{1mm}Varun Chandola\\
 \hspace{1mm}University at Buffalo, 338 Davis Hall\\
  Buffalo, New York 14260\\
}

\renewcommand{\shorttitle}{\textit{ChemTab} Deep Ensembles}
\date{}
\begin{document}
\maketitle

\begin{abstract}
Complete computation of turbulent combustion flow involves two separate steps: {\em mapping} reaction kinetics to low-dimensional manifolds and {\em looking-up} this approximate manifold during CFD run-time to estimate the thermo-chemical state variables. In our previous work, we showed that using a deep architecture to learn the two steps jointly, instead of separately, is 73 $\%$ more accurate at estimating the source energy, a key state variable, compared to benchmarks and can be integrated within a DNS turbulent combustion framework. In their natural form, such deep architectures do not allow for uncertainty quantification of the quantities of interest: the source energy and key species source terms. In this paper, we expand on such architectures, specifically \textit{ChemTab}, by introducing deep ensembles to approximate the posterior distribution of the quantities of interest. We investigate two strategies of creating these ensemble models: one that keeps the flamelet origin information ({\em Flamelets} strategy) and one that
ignores the origin and considers all the data independently ({\em Points} strategy). To train these models we used flamelet data generated by the GRI--Mech 3.0 methane mechanism, which consists of 53 chemical species and 325 reactions. Our results demonstrate that the \textit{Flamelets} strategy is superior in terms of the absolute prediction error for the quantities of interest, but is reliant on the types of flamelets used to train the ensemble. The \textit{Points} strategy is best at capturing the variability of the quantities of interest, independent of the flamelet types. We conclude that, overall, \textit{ChemTab Deep Ensembles} allows for a more accurate representation of the source energy and key species source terms, compared to the model without these modifications. 
\end{abstract}

\section{Introduction}
Accurate computation of turbulent combustion flow is necessary for the development of propulsion and energy generation applications (e.g., rocket motors \cite{dnssolid} \cite{ablate}, internal combustion engines \cite{iceles}, nuclear reactors \cite{nuclear}). Executing such simulations comes with a common, but critical choice: either to make simplifying assumptions \cite{GONZALEZJUEZ201726} (\textit{the simulation does not exactly model all the physics and chemistry, but is feasible}) or to use direct numerical simulations (DNS) which take long to run and consume large memory resources (\textit{simulation accurately models the phenomena, but is often infeasible}.) For example, a single DNS simulation of $CO/H_2/N_2$-air jet flames produces several terabytes of raw data and requires 120,000 hours of CPU-time \cite{Hawkes_2005}. Quantifying model and parameter uncertainty remains a challenge when representing such systems due to the different scales and non-linearity \cite{uqturb1,MUELLER2018137,JI20192175}.

Given the difference in scales of the flow and chemistry calculation,  the two sub--systems are typically (re)solved separately within a DNS framework. Approaches such as the {\em Flamelet Generated Manifolds} (FGM) use a two--step strategy where the governing reaction kinetics are pre--computed and mapped to a low--dimensional manifold, characterized by a few reaction progress variables (model reduction) and the manifold is then ``looked--up'' during CFD run--time to estimate the high--dimensional system state used by the flow system (Fig. \ref{fig:reduced}). Having these two separate calculations computed independently may reduce the estimation accuracy of the thermo-chemical state variables due to the assumption that a low-dimensional manifold exists and is an accurate mapping for the combustion chemistry \cite{maaspope92}. In previous work, we developed \textit{ChemTab} \cite{chemtab1.0}, a deep neural network architecture, that jointly performs the look-up process by learning the appropriate embedding of the reaction progress variables of laminar flames via an encoder and applying a deep regressor to approximate the manifolds and retrieve the thermo-chemical state variables. The joint process is 73$\%$ more accurate at estimating the source energy compared to benchmarks and is easily coupled within a DNS turbulent combustion framework. 

\begin{figure}[!h]
\begin{center}
  \include{figures/2step}
  \caption{(Re)solving flow and chemistry systems separately during a DNS.}
  \label{fig:reduced}
\end{center}
\end{figure}

\begin{table}[!h]
    \caption{Nomenclature}
    \label{tab:nomenclature}
    \centering
    \begin{tabular}{|c|c|}
    \hline
    FGM & Flamelet Generated Manifolds \\
    $\rho$ & Density of the mixture \\
    $\dot{S_i}$ & Source term of the $i^{th}$ species \\
    $S_e$ & Source energy \\
    $T$ & Temperature of the mixture \\
    $Y_i$ & Mass fraction of $i^{th}$ species \\
    $\mathcal{D}_i$ & Diffusivity of the $i^{th}$ species \\
    $k$ & Thermal conductivity \\
    $h_i$ & Enthalpy of the $i^{th}$ species \\
    $h_{f,i}^o$ & Heat of formation of the $i^{th}$ species \\
    $Z_{mix}$ & Mixture Fraction\\
    \hline
    \end{tabular}
\end{table}

While deep neural network architectures have, in general, become the go-to approach in tasks where the sole goal is accurate prediction, these models do not communicate any inherent uncertainty in their predictions. The task of manifold ``look-up'' within a turbulent combustion DNS is an approximation, which can lead to errors in the chemistry species calculations outside those used for training \cite{biogas}, leading to discrepancy between the approximate ``look-up'' model and truth, and is therefore a major source of uncertainty. Predictions from complex computational models with uncertainty quantification (UQ) increase trustworthiness, both in the model and underlying assumptions. 

The goal of this paper, is to expand the \textit{regressor} part of the \textit{ChemTab} deep architecture (Fig.  \ref{fig:chemtab-to-chemtab-deep-ensembles}) to be able to quantify the uncertainty in the quantities of interest (the species source energy $S_e$ and the key source terms $\dot{S}_{O2}, \dot{S}_{CO}, \dot{S}_{CO2}, \dot{S}_{H2O}, \dot{S}_{OH}, \dot{S}_{H2}, \dot{S}_{CH4}$) using Deep Ensembles (DE) \cite{deepensembles} and to study how two different data strategies for network training affect the response for the quantities of interest. The two strategies are to create training data ensembles keeping flamelet information (\textit{Flamelets} strategy) or to create training data ensembles without considering the underlying flamelet information (\textit{Points} strategy). The hypothesis is that \textit{Points} strategy results in models that, in general, predict the quantities of interest better, but \textit{Flamelets} strategy may be superior when predicting specific flamelet types.
\begin{figure}[!h]
\begin{center}
  \include{figures/chemtab-to-chemtab-deep-ensembles}
  \caption{\textit{ChemTab + Deep Ensembles} Architecture: Jointly generate embedded progress variables via linear dimensionality reduction (encoder) and appoximate the manifold via ensembled regressors to get the thermo-chemical state variables. The model predicts the key species source terms, which allow to calculate the source energy.}
  \label{fig:chemtab-to-chemtab-deep-ensembles}
\end{center}
\end{figure}

\subsection{\textit{ChemTab} Pre-training and Data Generation}
The governing conservation equations for mass, species, momentum and energy for the 1-D, fully compressible, and viscous flames, the steady-state solution with zero mass flux ($u_x = 0$) is given by:

\begin{equation}
\label{sample:problemform}
\pdv{}{x}(\mathcal{\rho} \mathcal{D}_i \pdv{Y_i}{x}) + \dot{S_i} = 0
\end{equation}
\begin{equation}
\label{sample:problemform2}
\pdv{}{x}(k \pdv{T}{x} + \sum{\mathcal{\rho}  \mathcal{D}_i \pdv{Y_i}{x} h_i}) - \sum{\dot{S_i} h_{f,i}^o }= 0
\end{equation}
Where $\rho$ is the density of the mixture, $\mathcal{D}_i$ is the diffusivity of the $i^{th}$ species, $Y_i$ is the $i^{th}$ species mass fraction, and $\dot{S_i}$ is the source term of the $i^{th}$ species. $k$ is the thermal conductivity, $T$ is the temperature of the mixture, $h_i$ is the enthalpy of the $i^{th}$ species, and $h_{f,i}^o$ is the heat of formation of the $i^{th}$ species. 

In Eq. ~\ref{sample:problemform2}, the final term $\sum{\dot{S_i} h_{f,i}^o }$ is the source energy $S_e$: the total sum of the product of all the source terms for all species and their respective heat of formation. The source terms and source energy are the quantities of interest (QoIs) in this problem. 

Solving these differential equations using a finite volume PDE solver provides the reaction progress variables. To model the chemical kinetics reaction rates, a variety of mechanisms are adopted in the combustion community. For this dataset, we used the GRI--Mech 3.0 methane mechanism, which consists of 53 chemical species and 325 reactions. The PDE flamelet solver discretizes the domain into $299$ grid points (299 observations on the axial coordinate $X_{pos}$) in between the fuel and the air boundary and $100$ flamelets were solved to steady--state. Once the solution reaches steady--state the solver completes one iteration. For the next iteration, the flame solution is strained by reducing the domain by $0.99$ and the process is continued until the flame extinguishes. Each flamelet is then tagged with the corresponding strain rate, the flamelet key. We thus received 29,900 data points (100 flames and 299 grid points) for a single pressure setting. Some of the generated data that represent extinguished flames (approximately 45 flames) were discarded, which led to exclusion of approximately 13,455 data points. So the final dataset that we used to train, test and validate the models consisted of 55 flamelets constituting 16,445 data points.

\newpage
\begin{subequations}
\label{eqn:datagenerated}
    \textbf{\text{Available Data from Chemistry Solver:}} $Y$ is a matrix of the $s$ species mass fractions, $\dot{S}$ is a matrix of the source terms of the $s$ species, $S_e$ is a vector of the source energy for the mixture, and $Z_{mix}$ is a vector of the mixture fraction. 
    \begin{alignat}{4}
        Y = \begin{bmatrix}
            Y_{11} &..&..&Y_{1s}\\
            ..&..&..&..\\
            ..&..&..&..\\
            ..&..&..&..\\
            Y_{n1}&..&.. & Y_{ns}
        \end{bmatrix},\quad
        \dot{S} = \begin{bmatrix}
            \dot{S}_{11} &..&..&\dot{S}_{1s}\\
            ..&..&..&..\\
            ..&..&..&..\\
            ..&..&..&..\\
            \dot{S}_{n1}&..&.. & \dot{S}_{ns}
        \end{bmatrix},\quad
        {S}_{e} = \begin{bmatrix}
            {S}_{e_{1}}\\
            ..\\
            ..\\
            ..\\
            {S}_{e_{n}}
        \end{bmatrix},\quad
        Z_{mix} = \begin{bmatrix}
            Z_{mix_{1}}\\
            ..\\
            ..\\
            ..\\
            Z_{mix_{n}}
        \end{bmatrix}
    \end{alignat}\\

    \textbf{\text{Embedded Data used by the Regressor(s) of \textit{ChemTab}}:} $Y$ is a matrix of the $s$ species mass fractions and $W$ is the weight matrix that encodes the original data into $p < < s$ dimensions.
    \begin{alignat}{3}
    \label{eqn:embeddingdatagenerated}
       C_{pv} =
       \begin{bmatrix}
            C_{pv_{11}} &..&C_{pv_{1p}}\\
            ..&..&..\\
            ..&..&..\\
            ..&..&..\\
            C_{pv_{11}} &..&C_{pv_{np}}
        \end{bmatrix} =
        Y \times W =
        \begin{bmatrix}
            Y_{11} &..&..&Y_{1s}\\
            ..&..&..&..\\
            ..&..&..&..\\
            ..&..&..&..\\
            Y_{n1}&..&.. & Y_{ns}
        \end{bmatrix}\quad \times \quad 
         \begin{bmatrix}
         W_{11} &..&W_{1p}\\
            ..&..&..\\
            ..&..&..\\
            ..&..&..\\
            W_{1s} &..&W_{sp}
        \end{bmatrix}
    \end{alignat}\\
    
    \textbf{\text{Predicted Quantities of Interest:}} The \textit{ChemTab} Regressor(s) use $C = C_{pv} \oplus Z_{mix}$ as inputs to predict the $k$ key source terms ($\hat{\dot{S}}$) and the source energy ($\hat{S_e}$).
    \begin{alignat}{3}
    \label{eqn:outputgenerated}
        \hat{\dot{S}} = \begin{bmatrix}
            \hat{S}_{11} &..&\hat{S}_{1k}\\
            ..&..&..\\
            ..&..&..\\
            ..&..&..\\
            \hat{S}_{n1} &..&\hat{S}_{nk}
        \end{bmatrix},\quad
        \hat{{S}_{e}} = \begin{bmatrix}
            \hat{{S}_{e_{1}}}\\
            ..\\
            ..\\
            ..\\
            \hat{{S}_{e_{n}}}
        \end{bmatrix},\quad
    \end{alignat}
\end{subequations}

In Eqs. \ref{eqn:datagenerated} the value of $n$ is 16,445 data points and the value of $s$ is 53. In Eq. \ref{eqn:embeddingdatagenerated} the value of $p$ is 4 and in Eq. \ref{eqn:outputgenerated} the value of $k$ is 7.

With the data generated, we previously trained the encoder part of \textit{ChemTab}, i.e., the embedding weights $W$ are pre-computed. Through this linear dimensionality reduction, we use the optimal reactive scalars for the embedding $C_{pv}$ together with the given mixture fraction $Z_{mix}$ to form the embedded progress variables, which are then the inputs to the deep ensemble regressors.

\subsection{Uncertainty Quantification in \textit{ChemTab}}

Quantifying uncertainty in deep neural networks is a topic of increasing interest in aerospace applications (e.g., Monte-Carlo dropout layers in a U-net for hybrid rocket regression rate estimation \cite{SURINA2022160}, deep ensembles for the inertia estimation of target and servicing spacecraft \cite{CHU2020106189}, and Bayesian networks for risk assessment of aerospace bolts \cite{aerospacebolts}). In Bayesian inference, predictive uncertainty is often decomposed into the distinct types of aleatoric and epistemic uncertainty. Aleatoric uncertainty refers to inherent uncertainty in the data, and is considered intractable and irreducible \cite{ABDAR2021243}.
Epistemic uncertainty refers to inadequate knowledge about the data distribution, and theoretically can be reduced by collecting more data or understanding the problem better. In the context of a deep network, however, these two traditional uncertainty definitions are not easily separable because any prediction from the trained model is impacted by the model form uncertainty, the aleatoric uncertainty in the data used for training, and the epistemic uncertainty in the input data of interest. 

In the hypothetical scenario where a fully Bayesian Neural Network could be deployed, the goal would be to exactly predict the posterior distribution of the source energy $S_e$ and the key source terms $\dot{S}_i$, expressed as $\mathcal{S}= \{S_e, \dot{S}_i \} $ :
\begin{equation}
\label{sample:posterior}
p(\mathcal{S}|C, D) = \int p(\mathcal{S}|C, \theta) p(\theta|D)d\theta
\end{equation}
Where $C$ is the input embedded progress variables ($C = C_{pv} \oplus Z_{mix} $), $\theta$ are the weights of the deep network, and $D$ is the training data. 

However, given that a fully Bayesian Neural Network is computationally infeasible  \cite{https://doi.org/10.48550/arxiv.1910.06539} to estimate the posterior distribution, we use Deep Ensembles as a Bayesian approximation \cite{https://doi.org/10.48550/arxiv.1912.02757}. Deep Ensembles refer to the technique of training a number of deep neural networks from ensembles of the available data (in this case the embedded progress variables). There are various ways of generating these ensembles: e.g., bagging/bootstrap aggregating or stacking \cite{ensemblecite}. For this case, given that the original \textit{ChemTab} model is low bias (i.e., no underlying assumptions about the form of the target function) and high variance (i.e., high likelihood to overfit the training set), bagging is the best option. We train homogeneous weak learners in different regimes of the available embedded flamelet progress variables and aggregate predictions from these weak learners such that the overall resulting output is more robust (i.e., lower variance than the original single learner.) In this case, the goal becomes to approximate the posterior distribution of the source energy $S_e$ and the key source terms $\dot{S}_i$, expressed as $\mathcal{S}= \{S_e, \dot{S}_i \} $ from the $N$ deep ensemble models:
\begin{equation}
\label{sample:bayesapprox}
p(\mathcal{S} | C,D) = \int p(\mathcal{S}|C, \theta) p(\theta|D)d\theta \approx \frac{1}{N} \sum_{i=1}^{N} p(\mathcal{S}|C, \theta_i)\,, ~~\theta_i \sim p(\theta|D)
\end{equation}
Where $C$ is the input embedded progress variables, $\theta_i$ are the weights of the $i^{th}$ ensembled network, and $D$ is the available training data.

To characterize the posterior distribution $p(\mathcal{S}|C, D)$, we use the mean of the predictions from the ensemble:
\begin{equation}
\label{sample:posterior_ensemble}
\mu_{\mathcal{S}|C, D} = \frac{1}{N} \sum_{i = 1}^N \hat{\mathcal{S}_i}
\end{equation}

And the sample standard deviation as the uncertainty in the predictions:
\begin{equation}
U_\mathcal{S} =\sqrt{\frac{\sum_{i = 1}^N (\hat{\mathcal{S}_i} - \mu_{\mathcal{S}|C, D})^2}{N-1}}
\end{equation}

\section{Problem Formulation and Ensemble Strategies}
Each of the deep regressor models trained from a data ensemble, can be expressed as the following layer-wise problem formulation:
\begin{equation}
f_{\theta_i}^{[0]} = C = C_{pv} \oplus Z_{mix} 
\end{equation}

The subsequent layers together make up the regressor that learns a non-linear function between the embedded progress variables and the quantities of interest (the source energy $S_e$ and the source terms $\dot{S}_i$).

\begin{equation}
f_{\theta_i}^{[l]} = \theta_i^{[l-1]}f_{\theta_i}^{[l-1]}+b_i^{[l-1]} \qquad \forall  \quad l   \in 2\leq l \leq L-1
\end{equation}
\begin{equation}
f_{\theta_i}^{[L]} = \sigma(\theta_i^{[L-1]}f_{\theta_i}^{[L-1]}+b_i^{[L-1]})
\end{equation}
Where , $\theta_i^{l}$ are the weights of the ensembled network $i$ at layer $l$, $b_i^{l}$ is the bias of the ensembled network $i$ at layer $l$, and $\sigma$ is the ReLU scalar activation function.

As mentioned before, we considered two separate strategies to train these ensembled deep networks: 
\begin{enumerate}
    \item \textbf{\textit{Flamelets} Strategy}: 80 \% of the available 55 flamelets (\textit{including all the data points belonging to those flamelets}) are used for \textbf{\textit{training--validation}} of the models and 20\% of the 55 flamelets are used as \textbf{\textit{holdout}} for the performance analysis.
    
    \item \textbf{\textit{Points} Strategy}: 80 \% of the available 16,445 data points (\textit{without considering which flamelets they belong to}) are used for \textbf{\textit{training--validation}} of the models and 20\% of the 16,445 data points are used as \textbf{\textit{holdout}} for the performance analysis.
\end{enumerate}
In each of these two strategies, each member of the ensemble is trained from randomly sampled data with replacement from the \textit{training--validation} set. Each sample includes 80 \% of the corresponding training set (i.e., the probability of a single data point or flame being included in a member model is 64 \%). For the performance analysis, the ensembles are then evaluated on the entire \textbf{\textit{holdout}} set.

\section{Comparison of Ensemble Strategies}

\subsection{Flamelets Strategy}

\subsubsection{Ensemble Size Ablation Study} 
As a first step, we conducted an ablation study to select the optimal number of member models in the ensemble. We investigated ensemble sizes $N$ between $ 2 \le N \le 12$.

\begin{figure}[!h]
\begin{center}
  \includegraphics[width=0.60\columnwidth]{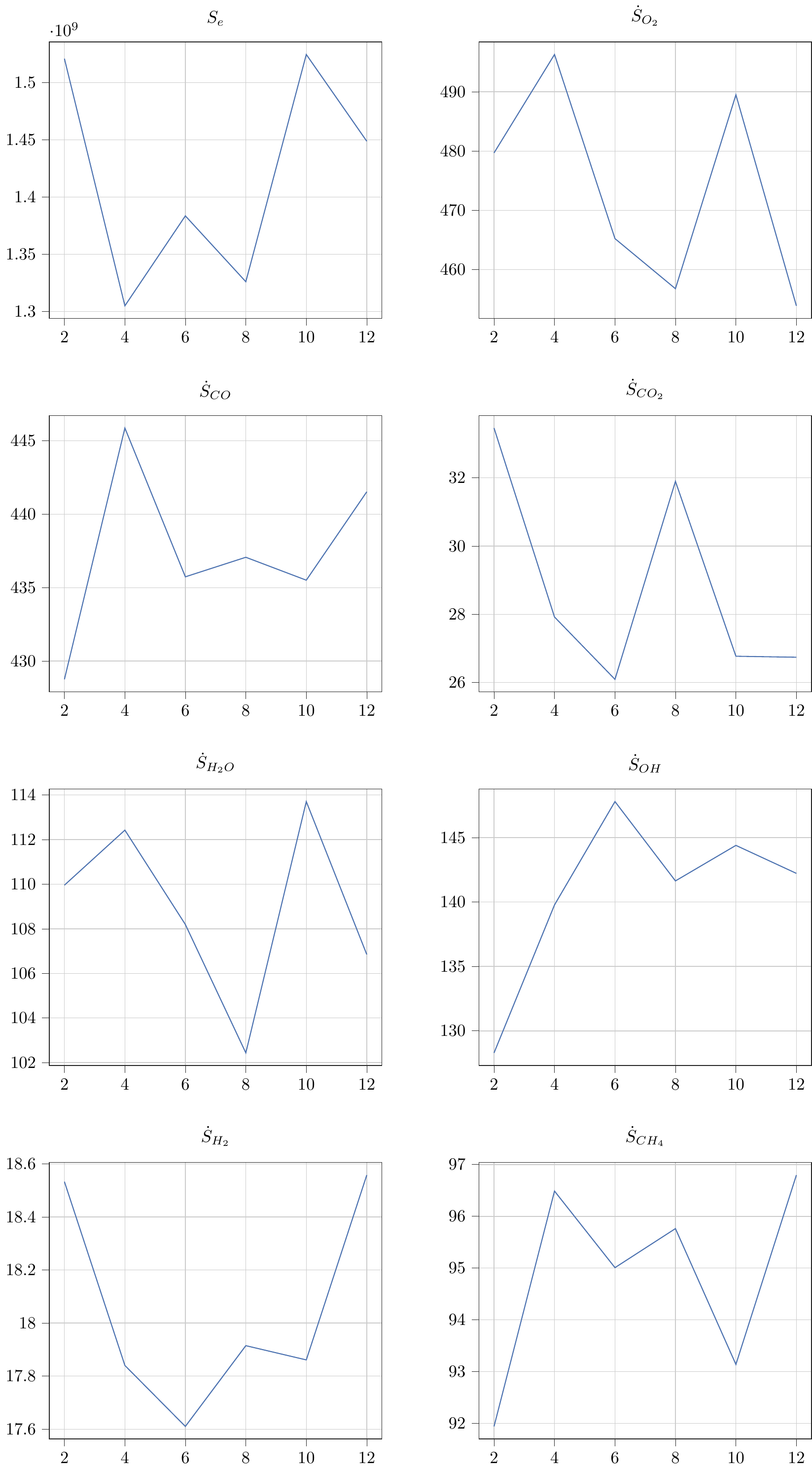}
  \caption{Flamelets Strategy - Ensemble Ablation}
  \label{fig:ensembleablationflames}
\end{center}
\end{figure}

The problem solved by the ensemble model here is highly non-linear and multi-objective and hence we do not see a clear optimal ensemble size that reduces the error consistently for all species source terms $\hat{\dot{S}}$ and source energy $\hat{S_e}$. Also, this strategy is highly biased towards the flamelets that were selected for the member models during the training--validation process. We observe, however, that the optimal membership size is between 6 to 10 for most of the source terms and source energy tests. We choose 8 as the optimal membership size for the performance analysis that follows.

\subsubsection{Performance Analysis}

To evaluate whether the ensembled models are adequate at predicting the quantities of interest, we compared the mean absolute prediction error (compared to truth) between the Single Model and Deep Ensembles architectures for the key source terms and source energy. All the comparisons are performed on the entirety of the \textit{holdout} dataset containing the 20\% flamelets that none of the models have been trained with. 

\begin{table}[!h]
    \caption{Mean Absolute Prediction Error by Key Source Terms}
    \label{tab:mae_sm_vs_ensemble_flames}
    \centering
    \begin{tabular}{|c|c|c|}
         \hline
        Term &  Single Model & Deep Ensemble \\
        \hline
       ${S}_{e}$ &          2.390641e+09 &           \textbf{1.325913e+09} \\
        $\dot{S}_{O_{2}}$ &          7.614887e+02 &           \textbf{4.567584e+02} \\
        $\dot{S}_{CO}$ &          5.408012e+02 &           \textbf{4.370647e+02} \\
        $\dot{S}_{CO_{2}}$ &          5.789108e+01 &           \textbf{3.189488e+01} \\
        $\dot{S}_{H_{2}O}$ &          1.792300e+02 &           \textbf{1.024356e+02} \\
        $\dot{S}_{OH}$ &          1.663385e+02 &           \textbf{1.416341e+02} \\
        $\dot{S}_{H_{2}}$ &          2.617136e+01 &           \textbf{1.791482e+01} \\
        $\dot{S}_{CH_{4}}$ &          1.906412e+02 &           \textbf{9.576332e+01} \\
        \hline
    \end{tabular}
\end{table}

When comparing by output quantity type, we observe in \ref{tab:mae_sm_vs_ensemble_flames} that the Deep Ensemble improves (decreases) the mean absolute error, compared to the single model, for all the quantities of interest.

\begin{table}[!h]
    \caption{Total Absolute Prediction Error on Source Energy (${S}_{e}$) by Flamelet Key}
    \label{tab:top10_flamelets_sm_vs_ensemble_flames}
    \centering
    \begin{tabular}{|c|c|c|}
         \hline
        Flamelet Key &  Single Model & Deep Ensemble \\
         \hline
          0.00024894 &              5.506834e+12 &               \textbf{1.908204e+12} \\
   0.00046070 &              7.157733e+11 &              \textbf{ 6.082280e+11} \\
   0.00059539 &              5.062102e+11 &             \textbf{  4.530053e+11} \\
   0.00065971 &              3.284762e+11 &              \textbf{ 2.980305e+11} \\
   0.00291989 &            \textbf{  1.514082e+11 }&               2.100692e+11 \\
   0.00085258 &             \textbf{ 1.443497e+11} &               2.128944e+11 \\
   0.00115982 &             \textbf{ 1.213801e+11} &               1.711747e+11 \\
   0.00174825 &            \textbf{  1.047254e+11} &               1.322080e+11 \\
   0.00193711 &             \textbf{ 1.004693e+11} &               1.289064e+11 \\
   0.00225936 &             \textbf{ 9.594018e+10} &               1.238106e+11 \\
    \hline
    \end{tabular}
\end{table}

When comparing by flamelet keys, we observe in Table  \ref{tab:top10_flamelets_sm_vs_ensemble_flames} that the Deep Ensemble improves (decreases) the residuals of the first few flames and then fails to improve on the subsequent flames. Given this result, it is evident that the ensemble models focus on reducing the absolute error of the flamelets that contribute the most absolute error, but fail to do so for the flamelets that contribute less. Another contributor to this result is that the prediction outputs from the models in the ensemble are equally weighted when taking a simple average. A different, chemistry-based weighting scheme, is likely more appropriate for this strategy. For example, if some members of the ensemble are better at predicting certain types of flamelets (because they were trained with such types of flamelets), then these members should be weighted more when predicting the quantities of interest for certain types of flamelets.

\begin{table}[!h]
    \caption{Total Absolute Prediction Error on (${S}_{e}$) by $X_{pos}$ Bins}
    \label{tab:top10_xpos_sm_vs_ensemble_flames}
    \centering
    \begin{tabular}{|c|c|c|}
         \hline
        $X_{pos}$ - Bins &  Single Model & Deep Ensemble \\
         \hline
          0.0 - 0.11 &              \textbf{9.220320e+11} &               1.202144e+12 \\
          0.11 - 0.22 &              1.680561e+12 &              \textbf{ 5.679709e+11} \\
          0.22 - 0.33 &              1.939765e+12 &             \textbf{  1.639737e+12} \\
0.33 - 0.44 &              1.695730e+12 &               \textbf{3.512237e+11} \\
0.44 - 0.55 &              4.271018e+11 &               \textbf{2.188931e+11} \\
0.55 - 0.66 &              4.738706e+11 &              \textbf{ 1.227035e+11} \\
0.66 - 0.77 &              4.508939e+11 &               \textbf{1.009125e+11} \\

0.77 - 0.88 &              2.092276e+11 &              \textbf{ 8.492034e+10} \\
0.88 - 0.99 &             \textbf{ 6.044873e+10} &               6.821203e+10 \\
 0.99 - 1.0 &              \textbf{3.187248e+09} &               4.210067e+09 \\
         \hline
    \end{tabular}
\end{table}

When comparing by grid position of the flamelet, we observe in Table  \ref{tab:top10_xpos_sm_vs_ensemble_flames} that the Deep Ensemble improves (decreases) the residuals for most of the $X_{pos}$ bins apart from the very beginning and end of the grid.

In the following pages, we show comparisons for 2 flamelets (Key 0.000024894 and 0.00046070) from the \textbf{\textit{holdout}} set between truth and the Single Model / Deep Ensemble. The 95\% confidence interval is defined as $1.96$ times the empirical standard deviation from the mean prediction. 

\begin{figure}[!h]
\begin{center}
  \includegraphics[width=1.0\columnwidth]{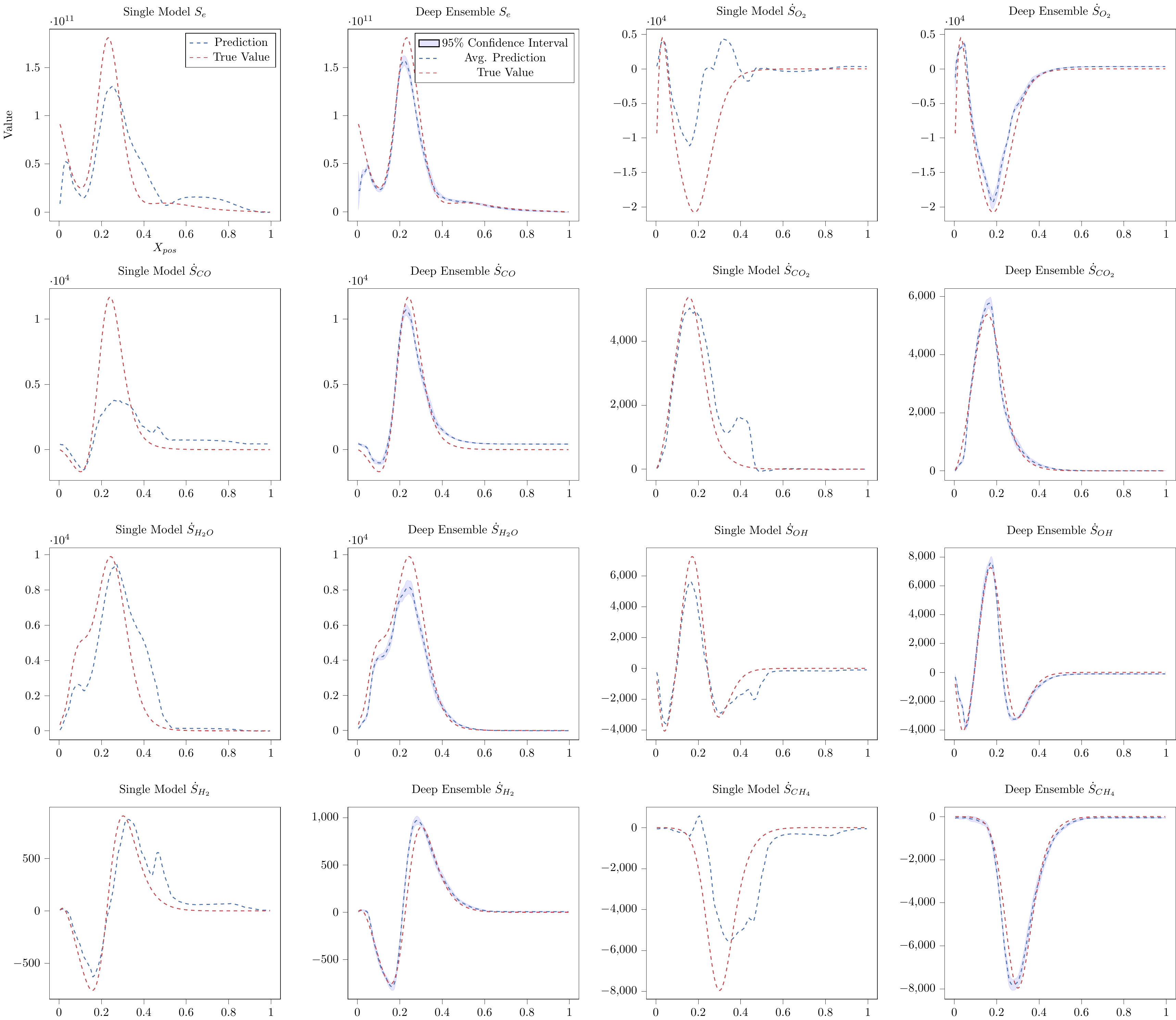}
  \caption{Single Model vs. Deep Ensemble comparison (Flamelet Key = 0.000024894)}
  \label{fig:flamelet_0_00024894_flames}
\end{center}
\end{figure}

We observe in Fig. \ref{fig:flamelet_0_00024894_flames} that the mean prediction of the ensemble is a much more accurate representation for the source terms and the source energy response. For example, for source terms $\dot{S}_{O_2}, \dot{S}_{CH_4}$, the single model appears to over and under estimate the true value, whereas the ensembled models capture the behavior smoothly. We also note that the 95\% confidence interval does not always fully encapsulate the true value.

\begin{figure}[!h]
\begin{center}
  \includegraphics[width=1.0\columnwidth]{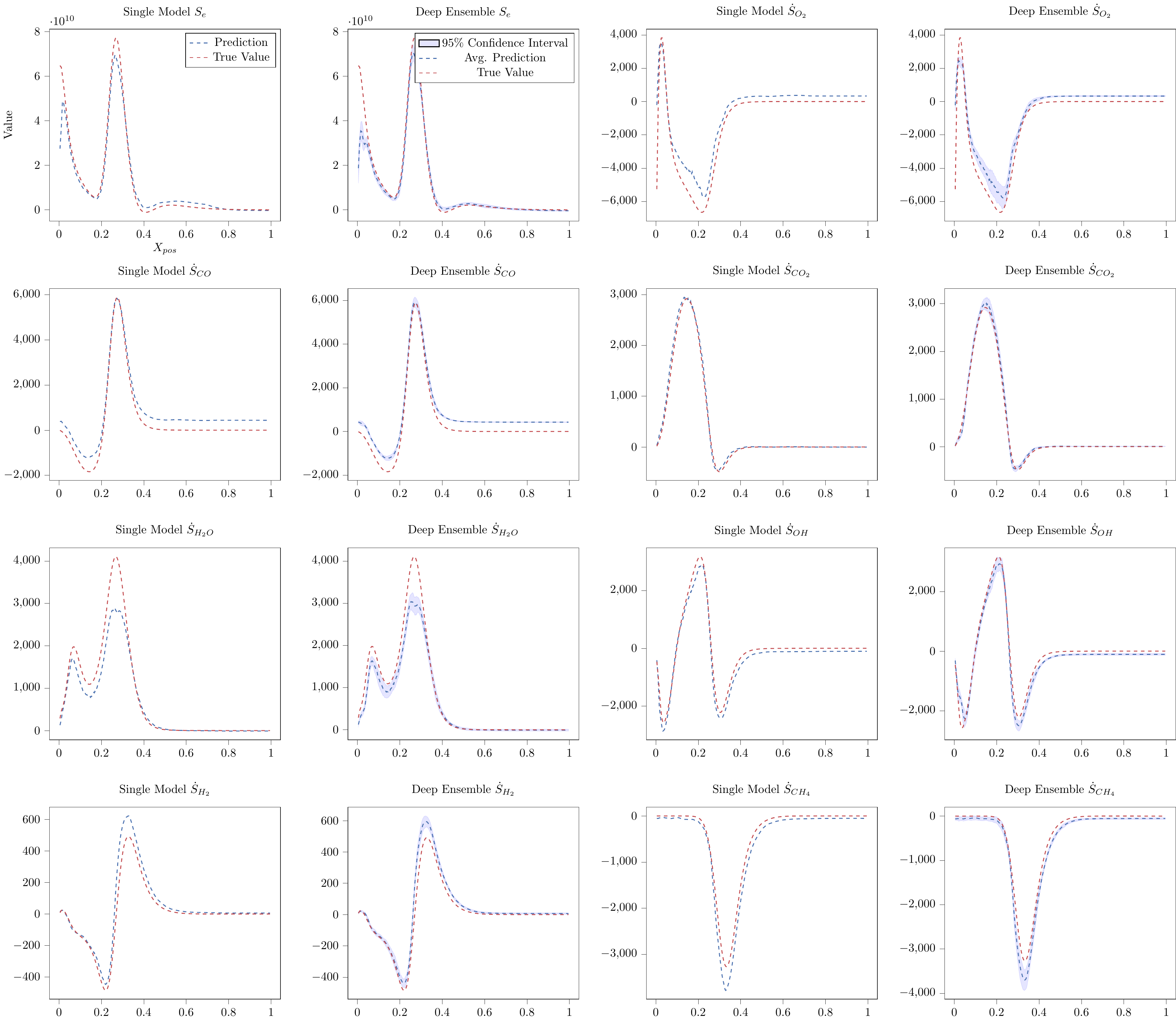}
  \caption{Single Model vs. Deep Ensemble comparison (Flamelet Key = 0.00046070)}
  \label{fig:flamelet_0_00046070_flames}
\end{center}
\end{figure}

We observe in Fig. \ref{fig:flamelet_0_00046070_flames} that the differences between the single and ensembled models are less pronounced that the previous flamelet. We note that both architectures do not capture the peaks in some of the source terms, e.g., $\dot{S}_{H_2O}$. It is, however, valuable that the ensembled models provide a uncertainty measure, which clearly communicates the existence of peaks or hills in the responses, e.g., for $\dot{S}_{O_2}, \dot{S}_{H_2O}$. From these predictions, we also observe the larger errors in the beginning portions of the flamelet for some quantities, e.g., in the source energy $S_e$ and $\dot{S}_{O_2}$ plots.

\subsection{Points Strategy}

\subsubsection{Ensemble Size Ablation Study} 
Similarly to the \textit{Flamelet} strategy, we conducted an ablation study to select the optimal number of member models in the ensemble for this case. We investigated ensemble sizes $N$ between $ 2 \le N \le 12$.

\begin{figure}[!h]
\begin{center}
  \includegraphics[width=0.60\columnwidth]{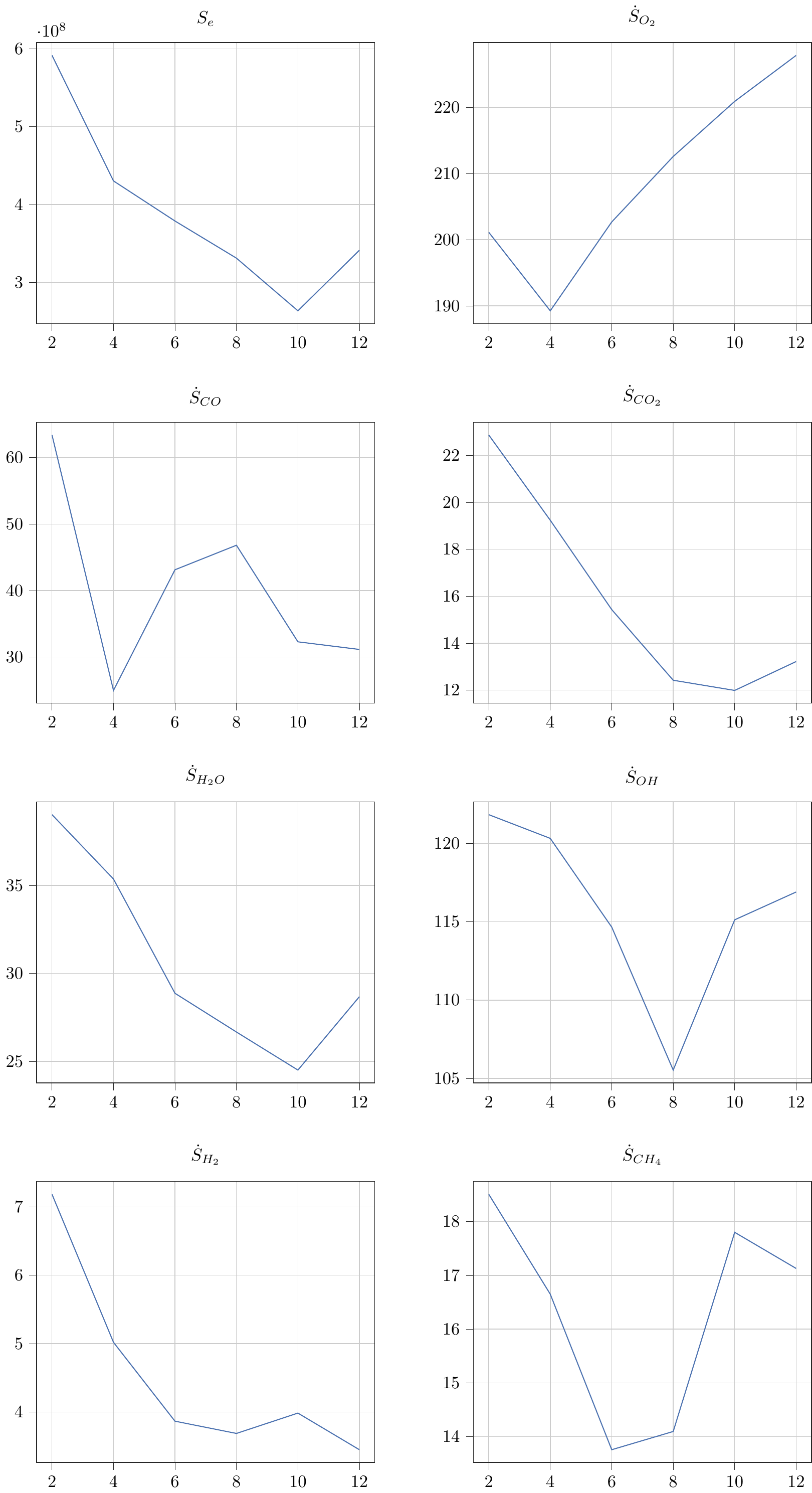}
  \caption{Points Strategy - Ensemble Ablation}
  \label{fig:ensemble_ablation_points}
\end{center}
\end{figure}

We observe in Fig. \ref{fig:ensemble_ablation_points} that several source terms show an optimal performance when the optimal membership size is 10. We choose 10 as the optimal membership size for the performance analysis that follows. 

\subsubsection{Performance Analysis}
To evaluate whether the ensembled models are better at predicting the quantities of interest, we compared the mean absolute prediction error (compared to truth) between the Single Model and Deep Ensembles architectures for the key source terms and source energy. All the comparisons are performed on the entirety of the \textit{holdout} dataset containing the 20\% data points that none of the models have been trained with. 

\begin{table}[!h]
    \caption{Mean Absolute Prediction Error by Key Source Terms}
    \label{tab:mae_sm_vs_ensemble_points}
    \centering
    \begin{tabular}{|c|c|c|}
         \hline
        Term &  Single Model & Deep Ensemble \\
         \hline
           ${S}_{e}$ &          4.679949e+08 &           \textbf{2.635308e+08} \\
         $\dot{S}_{O_{2}}$ &         \textbf{ 1.951080e+02 }&           2.208963e+02 \\
         $\dot{S}_{CO}$ &          3.353040e+01 &           \textbf{3.229429e+01} \\
        $\dot{S}_{CO_{2}}$ &          3.011369e+01 &           \textbf{1.199201e+01} \\
        $\dot{S}_{H_{2}O}$ &          3.879371e+01 &           \textbf{2.450645e+01} \\
         $\dot{S}_{OH}$ &          1.280294e+02 &           \textbf{1.151302e+02} \\
         $\dot{S}_{H_{2}}$ &          9.770577e+00 &           \textbf{3.982494e+00} \\
        $\dot{S}_{CH_{4}}$ &          4.241771e+01 &           \textbf{1.780133e+01} \\    
         \hline
    \end{tabular}
\end{table}

When comparing by output quantity type, we observe in Table \ref{tab:mae_sm_vs_ensemble_points} that the Deep Ensemble improves (decreases) the mean absolute error, compared to the single model, for all the quantities of interest except $\dot{S}_{O2}$.

\begin{table}[!h]
    \caption{Total Absolute Prediction Error on Source Energy (${S}_{e}$) by Flamelet Key}
    \label{tab:top10_flamelets_sm_vs_ensemble_points}
    \centering
    \begin{tabular}{|c|c|c|}
         \hline
        Flamelet Key &  Single Model & Deep Ensemble \\
         \hline
        0.00023650 & 7.904599e+10 &  \textbf{2.853646e+10} \\
        0.00020277 & 6.357353e+10 &  \textbf{3.267808e+10} \\
        0.00033866 & 6.025435e+10 &  \textbf{4.111402e+10} \\
        0.00030564 & 5.756831e+10 &  \textbf{3.318879e+10 }\\
        0.00022467 & 5.533021e+10 &  \textbf{3.334996e+10} \\
        0.00039499 & 5.167164e+10 &  \textbf{4.540711e+10} \\
        0.00032172 & 5.037682e+10 &  \textbf{3.497139e+10} \\
        0.00027584 & 4.817240e+10 &  \textbf{2.317479e+10} \\
        0.00024894 & 4.783312e+10 &  \textbf{2.263936e+10} \\
        0.00021344 & 4.749524e+10 &  \textbf{3.102894e+10} \\
    \hline
    \end{tabular}
\end{table}

When comparing by flamelet keys, we observe in Table \ref{tab:top10_flamelets_sm_vs_ensemble_points} that the Deep Ensemble improves (decreases) the residuals consistently across all flamelets, including the ones that previously the single model was better for. With this result, our initial hypothesis, that overall the \textit{Points} strategy will perform better than the \textit{Flamelets} strategy given the equal weighting of the ensemble models, is true.

\begin{table}[!h]
    \caption{Total Absolute Prediction Error on (${S}_{e}$) by $X_{pos}$ Bins }
    \label{tab:top10_xpos_sm_vs_ensemble_points}
    \centering
    \begin{tabular}{|c|c|c|}
         \hline
        $X_{pos}$ - Bins &  Single Model & Deep Ensemble \\
         \hline
          0.0 - 0.11 & 3.364565e+11 & \textbf{3.162390e+11} \\
          0.11 - 0.22 & 2.284694e+11 & \textbf{1.254176e+11} \\
         0.22 - 0.33 &  4.133929e+11 & \textbf{2.176143e+11} \\
           0.33 - 0.44 & 2.058881e+11 & \textbf{1.079951e+11} \\
           0.44 - 0.55 & 8.971007e+10 & \textbf{3.211109e+10} \\
        0.55 - 0.66 & 1.102133e+11 & \textbf{2.920763e+10} \\
        0.66 - 0.77 & 8.144498e+10 & \textbf{2.514758e+10} \\
        0.77 - 0.88 & 4.431664e+10 & \textbf{9.388437e+09} \\
        0.88 - 0.99 & 2.765116e+10 & \textbf{3.611705e+09} \\
         0.99 - 1.1 & 1.691913e+09 & \textbf{2.038095e+07} \\
         \hline
    \end{tabular}
\end{table}

When comparing by grid position of the flamelet, we observe in  \ref{tab:top10_xpos_sm_vs_ensemble_points} that the Deep Ensemble improves (decreases) the residuals forall of the $X_{pos}$ bins from the very beginning to the end of the grid.

In the following pages, we show comparisons for 2 flamelets (Key 0.00023650 and 0.00020277) from the \textbf{\textit{holdout}} set between truth and the Single Model / Deep Ensembles. The 95\% confidence interval is defined as $1.96$ times the empirical standard deviation from the mean prediction. 

\begin{figure}[!h]
\begin{center}
  \includegraphics[width=1.0\columnwidth]{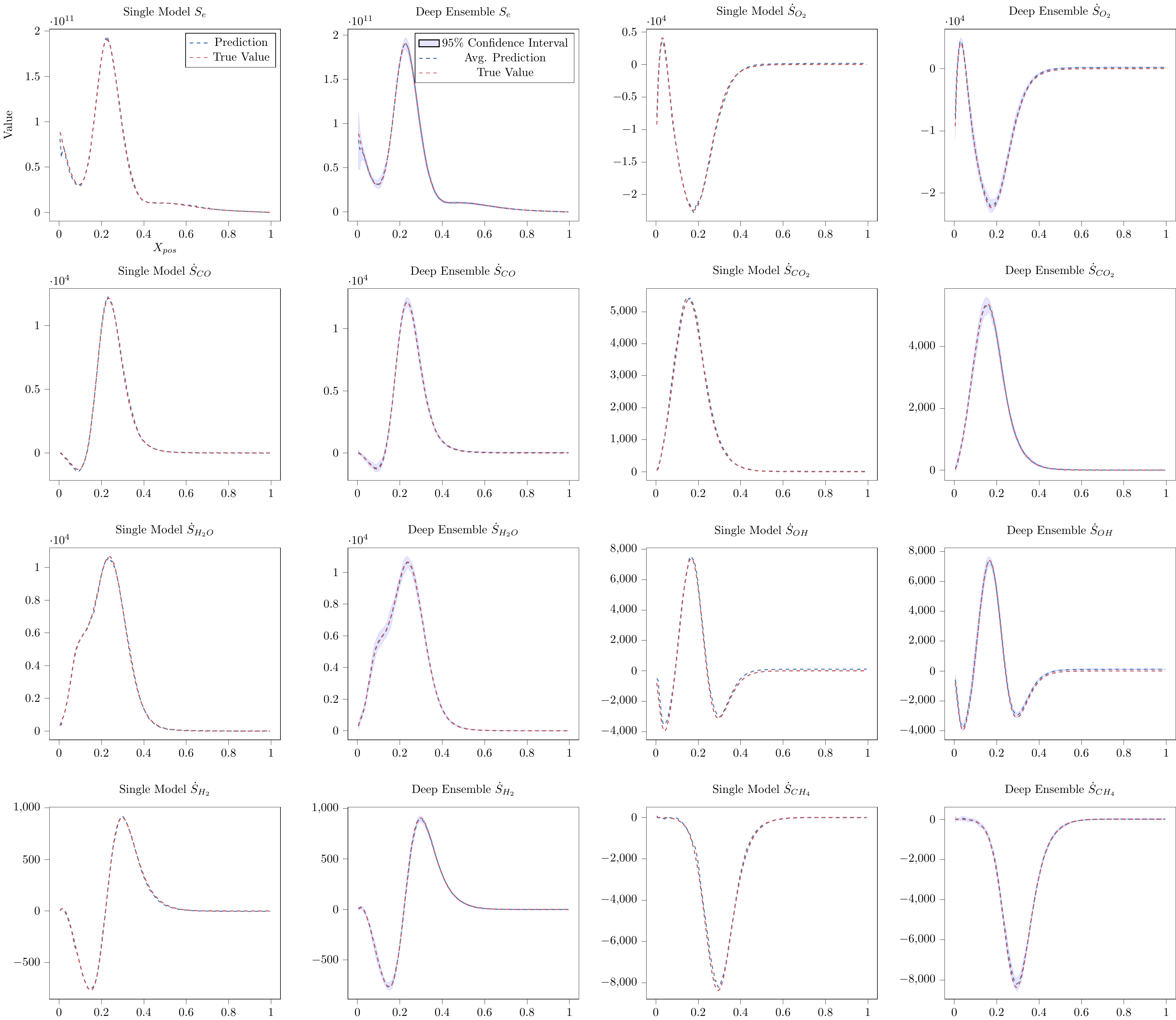}
  \caption{Single Model vs Deep Ensemble comparison (Flamelet Key = 0.00023650)}
  \label{fig:flamelet_0_00023650_points}
\end{center}
\end{figure}

Initially, we observe in Fig. \ref{fig:flamelet_0_00023650_points} that the \textit{Points} strategy, both for the single and for the ensembled models follows the quantities of interest closer compared to the \textit{Flamelets} strategy. Compared to the single model, 
the ensemble approach, offers the added advantage of knowledge where the model has higher uncertainty, which is visible for example in the plot for the source energy $S_e$ in the initial part of the flamelet. 

\begin{figure}[!h]
\begin{center}
  \includegraphics[width=1.0\columnwidth]{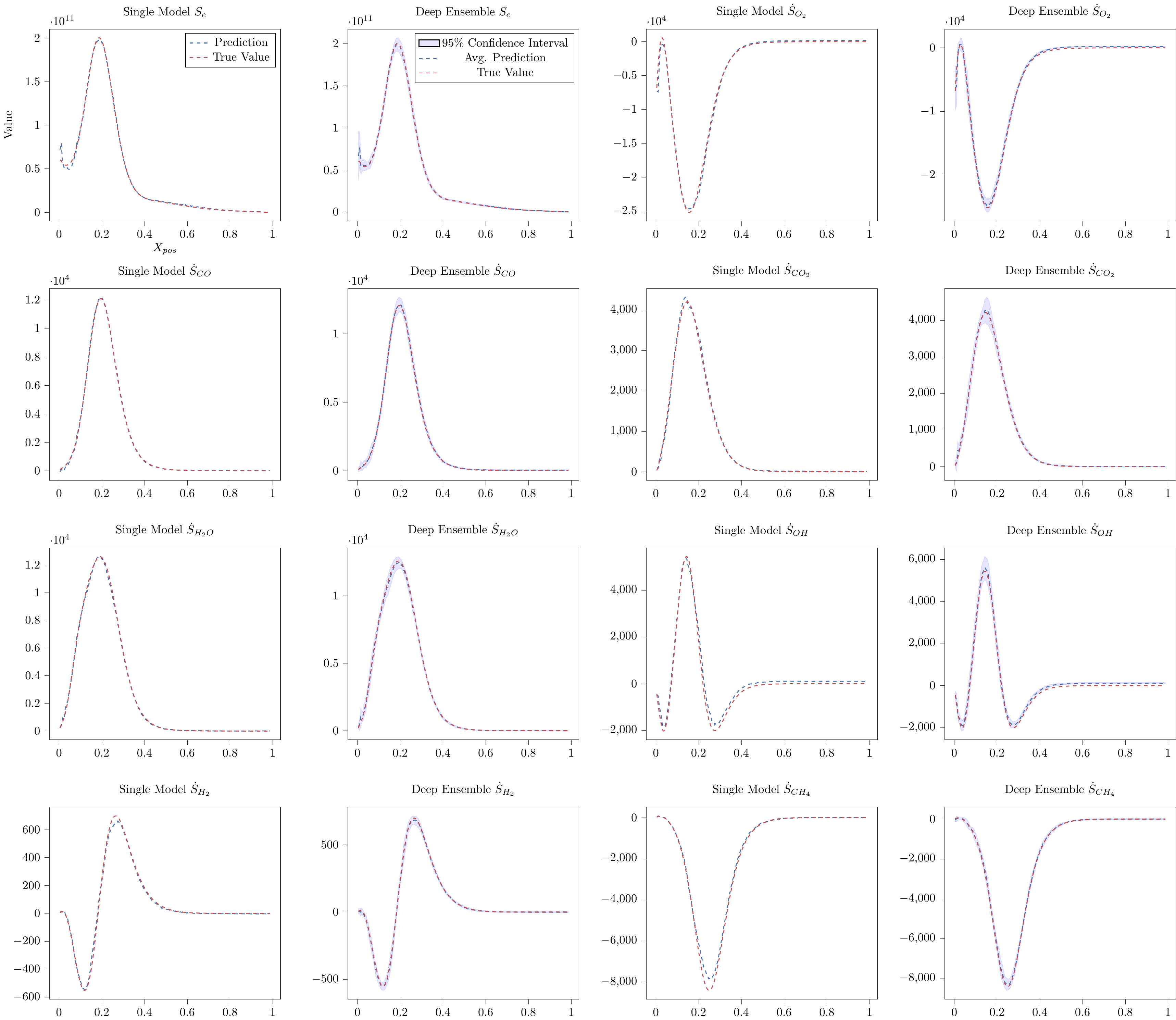}
  \caption{Single Model vs Deep Ensemble comparison (Flamelet Key = 0.00020277)}
  \label{fig:flamelet_0_00020277_points}
\end{center}
\end{figure}

For this flamelet, we observe in Fig. \ref{fig:flamelet_0_00020277_points} that the single model has some discrepancy and does not capture the variability for some of the source terms, e.g., $\dot{S}_{CH_4}$ in certain flamelet positions. However, the ensemble mean predictions not only captures that variability, but also includes the truth within the range of the confidence interval, which makes the predictions more trustworthy compared to the single model case.

\section{Conclusions}
In this paper we presented \textbf{\textit{ChemTab Deep Ensembles}}, a deep uncertainty-enabled architecture capable of predicting the key source terms and source energy from flamelet generated manifolds. We compared the mean absolute error for these quantities of interest with the original version of the architecture, which included a single regressor. We generated results from two strategies on creating the ensembles, one that keeps the flamelet origin information (\textit{Flamelets} strategy) and one that ignores the origin and considers all the data points independently (\textit{Points} strategy). Our results demonstrate that the (\textit{Flamelets} strategy) is superior in terms of the absolute prediction error for the quantities of interest, but is notably reliant on the types of flamelets used to train the ensemble. We noted that for certain flamelet keys and for certain positions within those flamelets (particularly the beginning or end), the (\textit{Flamelets} strategy) produced weaker results compared to the single model. In future work, we will explore a different, chemistry-based weighting scheme, for the \textit{Flamelets} strategy instead of averaging the ensemble model outputs, which may remedy these issues. The \textit{Points} strategy, as was our initial hypothesis, performs better in terms of absolute errors than the single model for every flamelet and position we tested on, as well as capturing the variability of the quantities of interest well. In future work, we will investigate methods to further capture the uncertainty throughout the deep architectures, including the encoder part.

\section*{Acknowledgments}
This work is supported by the United States Department of Energy’s (DoE) National Nuclear Security Administration (NNSA) under the Predictive Science Academic Alliance Program III (PSAAP III) at the University at Buffalo, under contract number DE-NA0003961.


\end{document}